\documentclass{article}

\usepackage[preprint]{neurips_2022}

\usepackage{amsmath,amsfonts,bm}

\def\eqref#1{equation~\ref{#1}}
\def\1{\bm{1}}

\DeclareMathAlphabet{\mathsfit}{\encodingdefault}{\sfdefault}{m}{sl}
\SetMathAlphabet{\mathsfit}{bold}{\encodingdefault}{\sfdefault}{bx}{n}

\usepackage[utf8]{inputenc} % allow utf-8 input
\usepackage[T1]{fontenc}    % use 8-bit T1 fonts
\usepackage{hyperref}       % hyperlinks
\usepackage{url}            % simple URL typesetting
\usepackage{booktabs}       % professional-quality tables
\usepackage{amsfonts}       % blackboard math symbols
\usepackage{nicefrac}       % compact symbols for 1/2, etc.
\usepackage{microtype}      % microtypography
\usepackage{xcolor}         % colors
\usepackage{subfig}

\usepackage{algpseudocode}
\usepackage{algorithm}
\usepackage{graphicx}

\DeclareUnicodeCharacter{2217}{*}

\title{Self-Programming Artificial Intelligence Using Code-Generating Language Models}

\author{%
  Alex Sheng \\
  New York University \\
  \texttt{alexsheng4@gmail.com} \\
  \And
  Shankar Padmanabhan \\
  University of Texas at Austin \\
  shankar.padmanabhan03@gmail.com\\
}

\begin{document}

\maketitle

\begin{abstract}
Recent progress in large-scale language models has enabled breakthroughs in previously intractable computer programming tasks. Prior work in meta-learning and neural architecture search has led to substantial successes across various task domains, spawning myriad approaches for algorithmically optimizing the design and learning dynamics of deep learning models. At the intersection of these research areas, we implement a code-generating language model with the ability to modify its own source code. Self-programming AI algorithms have been of interest since the dawn of AI itself. Although various theoretical formulations of generalized self-programming AI have been posed, no such system has been successfully implemented to date under real-world computational constraints. Applying AI-based code generation to AI itself, we develop and experimentally validate the first practical implementation of a self-programming AI system. We empirically show that a self-programming AI implemented using a code generation model can successfully modify its own source code to improve performance and program sub-models to perform auxiliary tasks. Our model can self-modify various properties including model architecture, computational capacity, and learning dynamics.
\end{abstract}

\section{Introduction}

Artificial intelligence technology has sought to emulate human cognitive capabilities within a computational substrate. Machine learning, and more recently deep learning, has enabled computers to derive programs from relevant data in order to address novel tasks for which they were not explicitly programmed.

Advanced artificial intelligence is theorized to require self-modifying programs that continuously imbue themselves with extended capabilities \citep{godel, schmidhuber:1987:srl}, surpassing their pre-programmed functionalities to achieve increased computational and predictive effectiveness. Meta-learning approaches have made significant progress along this avenue, from self-referential evolutionary algorithms \citep{schmidhuber:1987:srl} to RNN-based learners \citep{DBLP:journals/corr/SantoroBBWL16} and modern gradient-based meta-learners \citep{ DBLP:journals/corr/FinnAL17,adaptivemeta, DBLP:journals/corr/LiZCL17}. Adjacent work in neural architecture search \citep{amoebanet, hyperneat, nas} has led to the development of various algorithms for automatically optimizing the architectural parameters of neural networks. 

In recent years, AI-based computer code generation has seen dramatic progress \citep{codex, coderl, alphacode, codet5} stemming from breakthroughs in large-scale language processing models \citep{gpt3, bert, t5,  DBLP:journals/corr/VaswaniSPUJGKP17}. Modern language models (LMs) trained on computer programming tasks can not only generate syntactically correct computer code, but can also formulate logically correct computer programs from human-language prompts and understand pre-written programs to derive human-interpretable information.

At the intersection of these paradigms, we set out to explore a completely new avenue of research by combining self-modification approaches with code-generation approaches to achieve self-programming AI. Although theoretical formulations of self-programming AI \citep{godel} have been posed in the past, no practical implementation of a fully self-programming AI has been achieved to date under real-world computational constraints.

In this paper, we implement a LM-based code generation model with the ability to rewrite and improve its own source code, thereby achieving the first practical implementation of a self-programming AI system. With free-form source code modification, it is possible for our model to change its own model architecture, computational capacity, and learning dynamics. 

Since this system is designed for programming deep learning models, it is also capable of generating code for models other than itself. Such models can be seen as sub-models through which the main model indirectly performs auxiliary tasks. We explore this functionality in depth, showing that our model can easily be adapted to generate the source code of other neural networks to perform various computer vision tasks. We illustrate our system's ability to fluidly program other deep learning models, which can be extended to support model development in various other fields of machine learning.

\section{Related Work}

\textbf{Meta-learning.} Meta-learning is the subfield of machine learning that attempts to "learn how to learn". To be precise, let  $\omega$  denote an assumption on "how to learn" and $p(\mathcal{T})$ be a task distribution, where a "task" $\mathcal{T}=\{\mathcal{D}, \mathcal{L} \}$ for some dataset $\mathcal{D}$ and loss function $\mathcal{L}$ \citep{metaf}. Then, the goal of meta-learning can be viewed as computing $$\text{min}_{\omega}\mathbb{E}_{\mathcal{T}}(\mathcal{L}(\mathcal{D};\omega)).$$ Approaches in this subfield are broadly classified into RNN-based learners \citep{DBLP:journals/corr/SantoroBBWL16}, gradient-based meta-learners \citep{ DBLP:journals/corr/FinnAL17,adaptivemeta, DBLP:journals/corr/LiZCL17}, and self-referential evolutionary algorithms \citep{schmidhuber:1987:srl}. 

A select body of prior work within meta-learning establishes the conceptual groundwork for theoretical self-programming AI systems. Notably, Gödel machines \citep{godel} pose a theoretical formulation wherein an intelligent system can continuously rewrite itself in a globally optimal way. \citet{schmidhuber:1987:srl} discusses some of the evolutionary concepts that might underpin such systems.

\textbf{Automated Code Generation.} \citet{DBLP:journals/corr/VaswaniSPUJGKP17} introduced the transformer, an attention-based architecture that alleviated many of the previous issues with prior sequence-to-sequence models. The success of this architecture and subsequent innovations in Natural Language Processing \citep{gpt3, bert, t5}  have led to breakthroughs in AI-driven code generation \citep{codex, mathsolver, coderl, alphacode, codet5}.  

\citet{codex} revolutionized the application of language models to code generation. \citet{coderl} and \citet{codet5} extended the transfer learning concepts introduced in \citet{t5} to create systems capable of efficiently executing multiple tasks, including code generation and understanding. \citet{alphacode} created a system that was able to achieve top-half performances at various code competitions. \citet{mathsolver} showed that a pretrained language model fine-tuned on code was capable of solving, generating, and explaining university level math problems, as well as problems from challenging high-school math competitions. Although these systems are capable of writing programs for diverse and complex tasks, they are not capable of modifying and improving their own source code. 

\textbf{AutoML.} AutoML is a nascent and incredibly complex field that aims to automatically generate machine learning algorithms to solve problems without human tuning. Part of the difficulty lies in the fact that intricate, well-crafted neural networks are often necessary to obtain high performance on even mildly nontrivial problems such as CIFAR-100. Thus, the subfield of Neural Architecture Search(NAS) formed with the goal of creating algorithms capable of efficiently generating suitable architectures for a given problem. There are three approaches typically used for NAS: reinforcement learning-based NAS \citep{rlnassurvey, nas}, gradient-based NAS \citep{gradientnas}, and evolutionary NAS (ENAS) \citep{evo_nas}. 

Evolutionary algorithms are based upon the principles of Darwinian evolution and natural selection: a population of algorithms are created and evaluated, the "fittest" (best-performing) survive and are used as "parents" for a future generation of algorithms that are then evaluated, and so on \citep{evo_nas}. Some evolutionary approaches include regularized evolution search \citep{amoebanet, automl},  particle swarm optimization \citep{pso}, and genetic algorithms \citep{geneticalgo}. While AutoML techniques are usually restricted to one type of network (such as convolutional neural networks), our system is capable of editing and improving all sorts of neural network source code.

 \textbf{Our Contribution}. We extend the methodologies of automated code generation as well as some simple evolutionary ideas from the field of AutoML to create the first functional self-programming AI system. In particular, we use a code-generating transformer model based on \cite{t5} and a simple genetic algorithm \citep{geneticalgo}.

\section{Methods}
\subsection{Self-Programming AI}

\begin{figure}[]
    \centering
    \includegraphics[width=150mm]{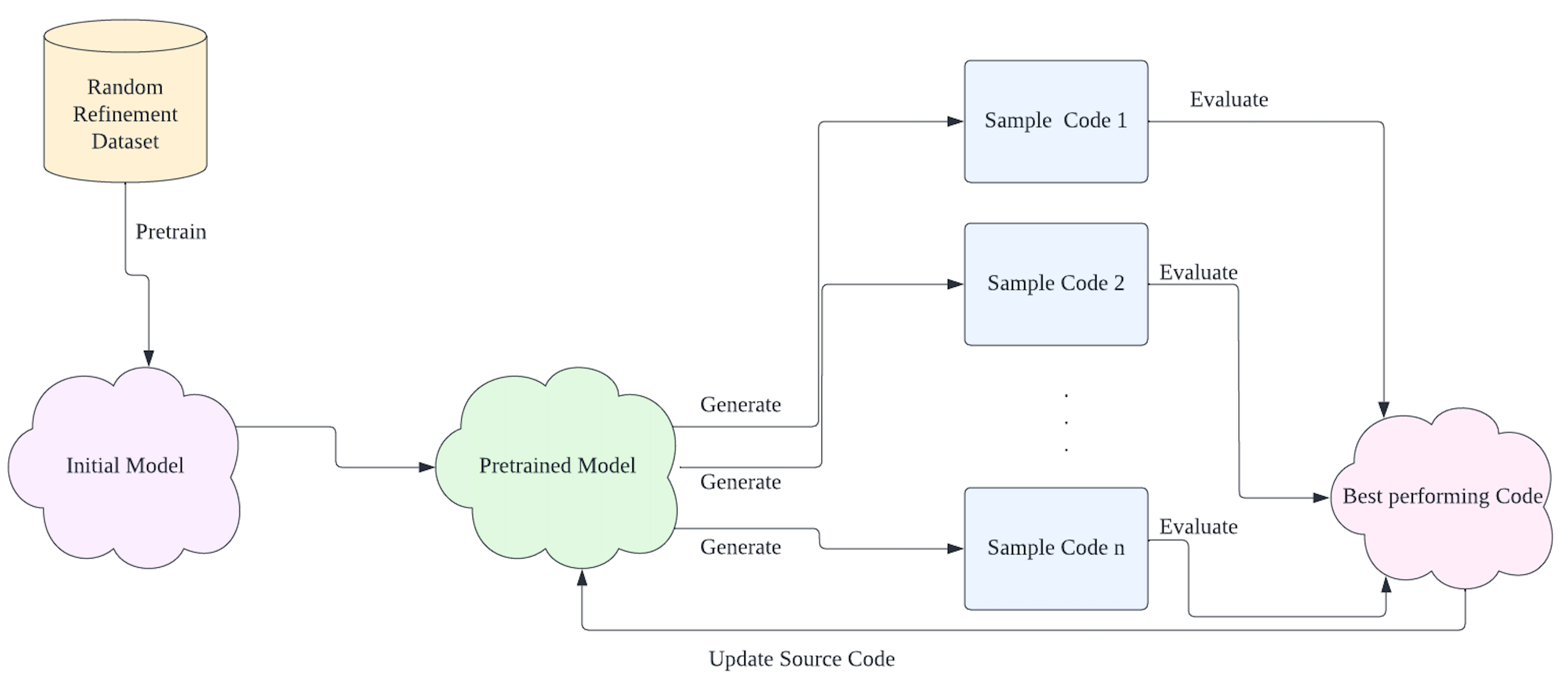}
    \caption{Baseline implementation of our model with a simple genetic algorithm.}
    \label{fig:base}
\end{figure}

We present the first practical implementation of a self-programming AI system. We initialize an arbitrary language processing model and train it on a synthetically-generated code corpus dataset, henceforth referred to as the "Random Refinement Dataset". Within this dataset, code samples are included for creating and training transformer-based neural networks. The model learns to parse and generate machine learning code, allowing it to understand its own source code. In an iterative process, the model is trained for a period of time before it is queried to generate a new source code for itself. A new iteration of the model is initialized and trained, and the cycle continues (Figure \ref{fig:base}).

Our model is initialized with a standard encoder-decoder transformer model based on T5 \citep{t5}. The model begins with 2 encoder layers, 2 decoder layers, 8 attention heads, and a feed-forward layer width of 1024. The architecture of the model is easily modified by its self-reprogramming in subsequent episodes.

Each episode of the algorithm is comprised of a training stage and a reprogramming stage. The model can be run for an arbitrary number of episodes to continuously self-improve, although its ability to do so is conditional on having an initial dataset with relevant code samples from which it can effectively deduce how to edit its existing source code.

During the training stage, the model is trained on the random refinement dataset. In our experiments, we train the model on a text-to-text "code refinement" task in order for it to learn code modification. Given an initial source code snippet, the model is trained to generate a modified version of that code snippet. The specific modification applied is arbitrary, but the newly generated or "refined" code should be syntactically and logically correct. The text-to-text transformer model outputs probability distributions over tokens, allowing for effective modeling of non-deterministic mappings for the code refinement process. 

The reprogramming stage utilizes an algorithm similar to the \textit{simple genetic algorithm} as described in \citet{geneticalgo}. During this stage, the trained model is given its own source code $c$ as input for code refinement. The model is queried to generate $n$ modified versions of its current source code $c_1, c_2, \cdots c_n$. \footnote{For most trials we used $n=8$. } Each refined source $c_i$ is executed to check for program validity, and the model instantiated within the refined code is quickly trained for one epoch on a subset $\mathcal{S}$ of the training corpus. Each model's average loss $\mathcal{L}_{\mathcal{S}}(c_i)$ over this "extended few-shot" training run is used as a proxy to evaluate the potential future performance of each refined source code version. The highest-performing source code $c_j$ is kept as the new source code for the model, and the training stage is repeated with the model defined in this source. To be precise, this $c_j$ satisfies $$ \mathcal{L}_{\mathcal{S}}(c_j)=\text{min}\left(\mathcal{L}_{\mathcal{S}}(c_1), \cdots \mathcal{L}_{\mathcal{S}}(c_n) \right).$$ 

\begin{algorithm}[hbt!]
\caption{Self-Reprogramming Algorithm}\label{alg:wordy}
\begin{algorithmic}
\Require{Default source code that defines a default model and default training algorithm}
\Ensure{Implements self-reprogramming artificial intelligence}
    \State $source \gets$ default source code;
    \While{Model is running}
        \If{$source$ defines a model}
            \State $model \gets$ model as defined in $source$;
        \EndIf
%        \If{$source$ defines a new dataset}
%            \State $dataset \gets$ dataset as defined in $source$;
%        \EndIf
        \If{$source$ defines a train() function}
            \State $train() \gets$ train() as defined in $source$;
        \EndIf
        \State $train(model, dataset)$;
        \For{number of candidates}
            \State $candidate \gets model.$generate$(source)$ 
            \State $model, train, dataset \gets$ as defined in $candidate$
            \State $candidate.$metrics$ = train(candidate)$ on a small subset of the dataset
        \EndFor
        \State $source \gets candidate$ with the best metrics;
    \EndWhile
    \label{algorithm:deepphoenix_algorithm}
\end{algorithmic}
\end{algorithm}

\subsection{Programming other AI}
The process to generate code for novel neural networks is similar. We initialize an encoder-decoder transformer based on T5 and train it on a random refinement dataset of code samples for arbitrary convolutional neural networks.  

Next, the model is queried to generate modifications of an initial source code snippet. In our experiments, this is a network with a single hidden layer of 16 neurons. The possible modifications include adding convolutional layers, changing the size of convolutional or hidden layers, and increasing the number of hidden layers. 

During the reprogramming stage, the trained model is given the default network as input for code refinement and is then queried to generate $n$ modified versions of this network. Each modification is first checked for program validity, and is then trained for two epochs on subset of data for a classic neural network problem (such as MNIST or CIFAR-10). The objective of each network is to minimize the cross-entropy loss, $$\mathcal{L}(\theta)=-\sum_{i=1}^n g_i \log(p_i),$$ where $\theta$ represents the model parameters, $n$ is the number of classes, $g_i$ is the ground-truth label, and $p_i$ is the softmax probability outputted for the $i$-th class. The best-performing algorithm is reused as input to the model for further code refinement, and the process continues. 
 \newline

\section{Experiments}
\subsection{Self-Programming AI}
We carry out preliminary experiments evaluating a basic implementation of self-reprogramming AI. We procedurally generate a code refinement dataset for transformer models, and use the our algorithm to train language processing models on this dataset in order to rewrite their own neural network source code.

In the random refinement dataset, we sample unrefined source codes for T5-based transformer models. Several model design parameters are randomly generated: number of encoder layers (between 1 and 8 inclusive), number of decoder layers (between 1 and 8), feed-forward layer dimensionality (between 64 and 4096), and number of attention heads (between 1 and 16). The key-value dimensionality is set to be the model dimensionality divided by the number of attention heads. 

We "refine" this source code by randomly modifying one of the model design parameters. Either one among the encoder depth, decoder depth, and number of heads is incremented or decremented, or the feed-forward width is randomly increased or decreased by a percentage of 1\% to 50\%, inclusive. The unrefined source code is used as input to the model, and the refined source is used as the corresponding output label.

\begin{figure}[!t]
    \centering
    \includegraphics[width=10cm]{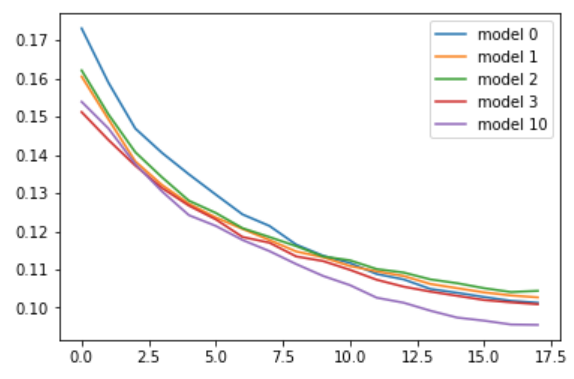}%
    \caption{Loss curves for text-to-text refinement}%
    \label{fig:loss}%
\end{figure}

In the training stage, the model is trained in a conditional generation configuration to generate refined transformer source codes given an unrefined source code as input. The objective of this procedure is to teach the model to modify design parameters in the source code of its own model class. This "refinement" process is more or less random, but the extended few-shot evaluation method used in the reprogramming stage allows us to achieve performance improvements over multiple iterations of our reprogrammed model.

In these experiments, we use a simple greedy search approach to evaluate and select generated source codes that would potentially improve the code refinement model. The model is queried with its own source code as input, and randomly generates 8 refined source code candidates. The candidates are each evaluated using the extended few-shot method, training for one epoch on the first 1024 samples of the code refinement dataset. 

Source code candidates that produce errors are discarded entirely, and the source code candidate with the lowest average training loss in extended few-shot evaluation is kept as the new query code. If all candidates underperform the original source code, then the original source is kept. This new query code is given to the code refinement model as input to generate another batch of source code candidates in the  following iteration of the main loop. The main loop runs for 10 iterations of training and reprogramming. The resulting loss curves of our self-reprogramming models are show in Figure \ref{fig:loss}, and an example of a modification made is in Table \ref{table:dataset-overview}.

\begin{table*}[phbt]
\setlength{\tabcolsep}{3pt}
\centering
\begin{tabular}{p{1.0\linewidth}}
\hline
\textbf{Unrefined Source (Input Text)} \\
\hline
config = transformers.PretrainedConfig.from\_pretrained("Salesforce/codet5-small") \\
config.num\_layers = 2 \\
config.num\_decoder\_layers = 2 \\
config.d\_ff = 1024 \\
model = transformers.T5ForConditionalGeneration(config) \\
\hline
\textbf{Refined Source (Output Text)} \\
\hline
config = transformers.PretrainedConfig.from\_pretrained("Salesforce/codet5-small") \\
config.num\_layers = 2 \\
config.num\_decoder\_layers = 2 \\
config.d\_ff = 1331 \\
model = transformers.T5ForConditionalGeneration(config) \\
\hline
\end{tabular}
\caption{Example of an input-output pair comprised of an unrefined transformer source code and a corresponding refined source code. In the refined source, the feed-forward layer dimensionality parameter is randomly increased by 30\%.}
\label{table:dataset-overview}
\end{table*}

\subsection{Programming other AI}
In the random refinement dataset, we sample unrefined source codes for convolutional neural networks. Several model design parameters are randomly generated: number of hidden layers $n$ (between 1 and 8 inclusive), number of convolutional layers $c$ (between 0 and 8), number of channels $h$ (between 16 and 512), and size of each hidden layer $s$ (between 16 and 1024). Any of these four hyperparameters is either incremented or decremented for each modification.

The remainder of the process is similar to the self-programming stage. The model is given a initial source code and queried to generate $n$ modified versions. Each modified version is then trained a subset of the target dataset (such as CIFAR-10 or MNIST) before being evaluated on a disjoint subset of the target dataset. The source code with the highest accuracy is then given to the model as a new source, and the process is repeated over the next five main loop iterations. 

The performance of the code generated was evaluated on a handful of classic neural network tasks - MNIST, CIFAR10, and EMNIST (ByClass). Each experiment was repeated multiple times with different model setups.
\begin{figure}[]
    \centering
    \subfloat[\centering CIFAR-10]{{\includegraphics[width=6.5cm]{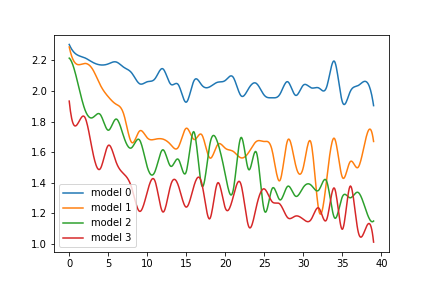} }}%
    \qquad
    \subfloat[\centering EMNIST]{{\includegraphics[width=6.5cm]{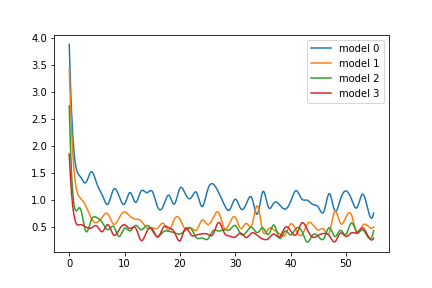} }}%
    \caption{Loss curves of multiple iterations of self-reprogramming models on convolutional networks for CIFAR-10 and EMNIST. Each graph represented Model 0 refers to the default model. The value of $n$ is the number of candidates tested prior to source code updating. Loss curves for MNIST do not show much improvement as even the input network was able to achieve high performance.}%
    \label{fig:geneticloss}%
\end{figure}

The accuracy of generated networks trained on MNIST increased from roughly 92\% to 98\% within one training epoch. Accuracy on CIFAR10 improves from 20\% to 70\% over four epochs. Finally, accuracy on EMNIST increased from roughly 70\% to 87\% over four epochs. An example of a modification made by the agent is provided in Table \ref{table:automl-change}. As shown in Figure \ref{fig:geneticloss}, later models tend to achieve favorable start, end, and intermediate loss values compared to earlier iterations.

\begin{table*}[phbt]
\setlength{\tabcolsep}{3pt}
\centering
\begin{tabular}{p{1.0\linewidth}}
\hline
\textbf{Input Network} \\
\hline 
  Linear(in=784, out=64, bias=True) \\
  ReLU() \\
  Linear(in=64, out=64, bias=True) \\
  ReLU() \\
  Linear(in=64, out=10, bias=True) \\
\hline
\textbf{Output Network} \\
\hline
  Conv2d(1, 128, kernel=(3, 3), stride=(1, 1)) \\
  Conv2d(128, 256, kernel=(3, 3), stride=(1, 1)) \\
  MaxPool2d(kernel=2, stride=2, padding=0, dilation=1) \\
  Linear(in=6400, out=64, bias=True) \\
  ReLU() \\
  Linear(in=64, out=64, bias=True) \\
  ReLU() \\
  Linear(in=64, out=62, bias=True)  \\
\hline
\end{tabular}
\caption{Example of an input neural network and code modifications made by our model. Our system adds two convolutional layers, a pooling layer, and automatically adjusts the number of inputs to the dense layers accordingly.}
\label{table:automl-change}
\end{table*}

\section{Analysis}
Our self-programming results verify that our model is capable of editing its own source code to extend its capabilities.  

Furthermore, on all three tasks for programming other models, our system is able to autonomously design networks that achieve effective performance. Our approach is both flexible by way of its free-form programming capabilities and simple in its search and selection procedures. Other methods to search for optimal networks, such as reinforcement learning, may be unstable \citep{rlunstable} when applied to different domains, and therefore may require specialized tuning in order to be effective.

Accuracy on MNIST is around 1.5\% below SoTA and EMNIST ByClass is about 10\% below SoTA, but performance on CIFAR-10 lags significantly behind the current SoTA \citep{sotas}. This is likely due to the fact that competitive results on CIFAR-10 typically require specialized model designs, which are unlikely to be achieved since our approach uses a simple class of models with greedy search over random code modifications. Performance on these three problems could possibly be improved to by using more sophisticated source code data and search algorithms specialized for computer vision model development.

\begin{figure}[!t]
    \centering
    \subfloat[\centering Total Candidates]{{\includegraphics[width=6.5cm]{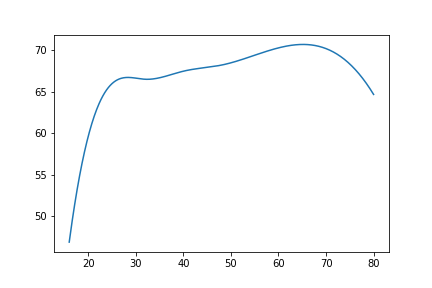} }}%
    \qquad
    \subfloat[\centering Epochs vs Candidates]{{\includegraphics[width=6.5cm]{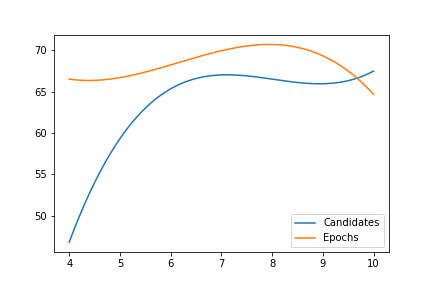} }}%
    \caption{The first graph (left) illustrates the average performance of a network on CIFAR-10 based on the total number of candidates tested over all epochs. The second (right) shows the average performance of a model with a varying number of candidates over four epochs and the average performance of a model with 8 candidates over a varying number of epochs.}%
    \label{fig:candidates}%
\end{figure}

As shown in Figure \ref{fig:candidates}a, the average performance of generated models on CIFAR-10 tends to increase with the total number of candidates (the product of the number of code search iterations multiplied by the number of candidates per iteration) up to a certain point. Figure \ref{fig:candidates}b shows that the average performance of generated models increases significantly with the number of candidates tried per epoch, and is not as sensitive to the number of code search iterations.

\begin{figure}[hbt!]
    \centering
    \subfloat[\centering EMNIST]{{\includegraphics[width=6.5cm]{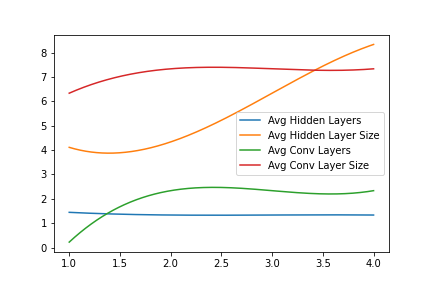} }}%
    \qquad
    \subfloat[\centering MNIST]{{\includegraphics[width=6.5cm]{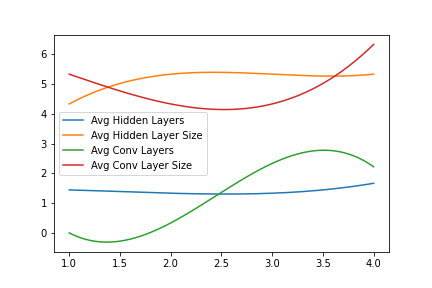} }}%
    \qquad
    \subfloat[\centering CIFAR]{{\includegraphics[width=6.7cm]{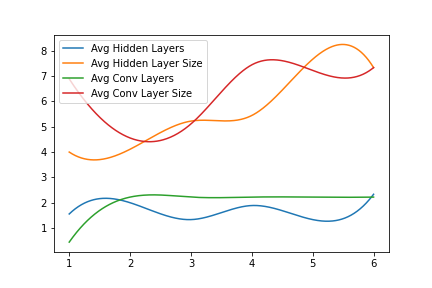} }}
    \caption{Average values of model design parameters per epoch for convolutional networks on EMNIST, MNIST, and CIFAR. The "avg hidden layer size" and "avg conv layer size" are represented as $\log_2$ of actual values. The results for MNIST and EMNIST are shown for the first four epochs, because neither network performance nor model design parameters changed significantly in subsequent epochs.}%
    \label{fig:hyper}%
\end{figure}

Figure \ref{fig:hyper} plots the average model size (number of parameters) of generated models over 10 consecutive reprogramming steps. The most significant increases were found to typically occur in the number of convolutional layers. Other significant increases occurred in the size of the hidden layers. The size of the convolutional layers did not increase significantly in any of the three tasks (it was found to fluctuate for CIFAR-10 and MNIST), and the number of hidden layers was rarely found to be increased. 

In the three tasks involving generating other models, the code-generating model consistently chose to increase convolutional layer counts, which was found to produce significant increases in performance. These deeper convolutional network source codes were naturally reused by the algorithm as input in subsequent code generation queries. In the initial training loop, most networks had no convolutional layers and a small minority had a single convolutional layer. Within four loops, every network had at least 2 convolutional layers and often more. The number of hidden layers did not significantly increase, indicating that deeper feed-forward architectures may not be necessary to achieve increased performance on these problems.

\begin{figure}[!t]
    \centering
    \includegraphics[width=9cm]{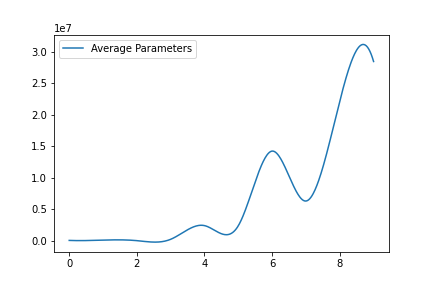}%
    \caption{Average number of parameters in best-performing model over 10 epochs. Analysis was conducted on CIFAR-10. Networks in the other two tasks converged in four or fewer epochs.}%
    \label{fig:avgparams}%
\end{figure}

As can be seen in \ref{fig:avgparams}, the number of trainable parameters in the best-performing network increased significantly over time. This agrees with our current understanding of neural networks that larger models with more parameters are more expressive and capable of effectively addressing more complex tasks \citep{scaling2, scaling}.

All in all, our results demonstrate that our system is able to effectively rewrite the code for a variety of types of neural networks. From a self-programming perspective, our self-programming AI system is capable of extending itself with additional functionalities by way of programming sub-models that can perform auxiliary tasks. From a more practical perspective, our neural network code generation approach can be seen as a kind of free-form AutoML solution that can be used to support model development in various task domains.

\section{Conclusion}
We propose and experimentally validate the first functional self-reprogramming AI system. In our experiments, we successfully implement the system and show it can freely modify its own neural network design. Over several iterations of the algorithm, reprogrammed models tend to achieve increasingly favorable performance on a simple supervised text-to-text code modification task.

Furthermore, the code-generating model used in our self-programming system can easily be adapted to program other models, in order to perform various other machine learning tasks.

\subsection{Future Work}

Our work presents a number of promising directions for future research. We are interested in extending this work by incorporating training on a large computer code corpus. Not only would this allow our model to adapt to differences in code styles (such as syntax differences or distinctions between machine learning libraries and frameworks), but it would also allow the model to incorporate code and synthesize new approaches to improve itself and its auxiliary sub-models.

Furthermore, an actor-critic implementation in which the code generation model is trained as an actor via gradients from an associated critic model, similarly to CodeRL in \citep{coderl}, could improve performance and generalization. We plan to implement these ideas promptly in future work. 
\newline \text{} 
\newline
\small

\bibliographystyle{plainnat}
\bibliography{references}

\end{document}